\documentclass[letterpaper, 10pt, conference]{ieeeconf}      

\IEEEoverridecommandlockouts                              

\usepackage{cite}
\usepackage{color}

\usepackage{graphicx}
\usepackage[font={small}]{caption}
\usepackage{amsmath}
\usepackage{amssymb}

\usepackage{tikz}
\usetikzlibrary{arrows} 
\usetikzlibrary{calc} 
\usepackage{pgfplots}

\usepackage{comment}

\newtheorem{remark}{Remark}

\title{\LARGE \bf
Path Planning for Autonomous Bus Driving in Urban Environments
}

\author{Rui Oliveira$^{1, 2}$, Pedro F. Lima$^{2}$, Gon\c{c}alo Collares Pereira$^{1, 2}$, Jonas M{\aa}rtensson$^{1}$ and Bo Wahlberg$^{1}$
\thanks{*This work was partially supported by the Wallenberg AI, Autonomous Systems and Software Program (WASP) funded by the Knut and Alice Wallenberg Foundation.}
\thanks{$^{1}$Department of Automatic Control, School of Electrical Engineering and Computer Science, KTH Royal Institute of Technology, Stockholm, Sweden
        {\tt\small rfoli@kth.se}, {\tt\small gpcp@kth.se}, {\tt\small jonas1@kth.se}, {\tt\small bo@kth.se}}%
\thanks{$^{2}$Scania, Autonomous Transport Solutions, S{\"o}dert{\"a}lje, Sweden 
		{\tt\small pedro.lima@scania.com} }%
}

\begin{document}

\maketitle
\thispagestyle{empty}
\pagestyle{empty}

\def\deltaS{ \Delta s}
\def\steeringAngleSymbol{ \phi }

\def\edgeConstantTermS{\bar{s}^{\text{edge}}}
\def\edgeConstantTermEy{\bar{e}_y^{s_k}}
\def\linearizationRefS{\bar{s}}
\def\linearizationRefEy{\bar{e}_y}
\def\linearizationRefEpsi{\bar{e}_\psi}

\def\CartesianPosition{\mathbf{p}}
\def\CartesianPositionConverted{\hat{\CartesianPosition}}
\def\SConverted{\hat{s}}
\def\EYConverted{\hat{e}_y}

\def\arcCircleRadius{r_{\pm}}

\def\geometricTransform{\mathcal{T}}
\def\geometricTransformEY{\geometricTransform_{e_y}}

\def\realNumbers{\mathbb{R}}


\def\optimizationVariablesEy{ e_y }
\def\optimizationVariablesEpsi{ e_\psi }
\def\optimizationVariableU{ u }

\def\DiscretizationSamplingDistance{\Delta s}
\def\linearizationReferenceVariablesS{ \mathbf{\bar{s}} }
\def\linearizationReferenceVariablesEy{ \mathbf{\bar{e}_y} }
\def\linearizationReferenceVariablesEpsi{ \mathbf{\bar{e}_\psi} }
\def\linearizationReferenceVariablesU{ \mathbf{\bar{u}} }

\def\solutionVariablesEy{  \mathbf{e}^*_y }
\def\solutionVariablesEpsi{  \mathbf{e}^*_\psi }
\def\solutionVariableU{  \mathbf{u}^* }

\def\optObjCenterDriving{ J_{ \text{center} } }
\def\optObjOverhang{ J_{ \text{overhang} } }
\def\optObjSmooth{ J_{ \text{smooth} } }

\def\optConstraintObsMatrix{ P_{i} }
\def\optConstraintObsOffset{ p_{i} }
\def\optConstraintObsPosition{ p_{e_y}^{\text{obs},i} }
\def\optConstraintAxleMatrix{ Q_{i} }
\def\optConstraintAxleOffset{ q_{i} }
\def\optConstraintRoadPosition{ q_{e_y}^{\text{driv},i} }
\def\optConstraintOverhangMatrix{ R_{i} }
\def\optConstraintOverhangOffset{ r_{i} }
\def\optConstraintOverhangRoadPosition{ r_{e_y}^{\text{driv},i} }
\def\optConstraintOverhangSlack{ \sigma_{i}^{r} }
\def\optVariableSigmaOverhang{ \sigma_{\text{oh}} }

\def\optConstraintMaxU{ u_{\max} }
\def\optConstraintMaxUDot{ {u}_{\max}' }
\def\optConstraintMaxUDotDot{ {u}_{\max}'' }
\def\optConstraintFirstDifferenceMatrix{ \mathbf{D_1} }
\def\optConstraintSecondDifferenceMatrix{ \mathbf{D_2} }

\def\offlinearizationToleranceEy{ \epsilon_y }
\def\offlinearizationToleranceEpsi{ \epsilon_\psi }


\def\maxRoadCurvature{ \kappa_{\text{road}} }
\def\maxBusCurvature{ \kappa_{\max} }

\def\VectorObstacleAvoidanceConstraint{v_i}
\def\ConstantObstacleAvoidanceConstraint{k_i}

\begin{abstract}
Driving in urban environments often presents difficult situations that require expert maneuvering of a vehicle.
These situations become even more challenging when considering large vehicles, such as buses.
We present a path planning framework that addresses the demanding driving task of buses in urban areas.
The approach is formulated as an optimization problem using the road-aligned vehicle model.
The road-aligned frame introduces a distortion on the vehicle body and obstacles, motivating the development of novel approximations that capture this distortion.
These approximations allow for the formulation of safe and non-conservative collision avoidance constraints.
Unlike other path planning approaches, our method exploits curbs and other sweepable regions, which a bus must often sweep over in order to manage certain maneuvers.
Furthermore, it takes full advantage of the particular characteristics of buses, namely the overhangs, an elevated part of the vehicle chassis, that can sweep over curbs.
Simulations are presented, showing the applicability and benefits of the proposed method.

\end{abstract}

\section{Introduction}

Autonomous driving is a fast developing technology, with promises of increased safety, efficiency, and comfort.
One interesting application is the automation of public urban transportation systems~\cite{Bishop:2000:Survey}.
Urban driving presents several challenges, such as tight roads and complex maneuvers, which become even harder when considering buses.

Path planning deals with the problem of finding paths for a vehicle to follow.
These paths must be collision free and comply with the non-holonomic constraints of the vehicle.
Furthermore, they should be optimized with respect to criteria such as comfort, safety, and efficiency.
Path planning is the subject of extensive research, however, a literature review shows a lack of applications targeting buses (mostly passenger vehicles are considered, as indicated by surveys~\cite{Katrakazas:2015:Survey, Frazzoli:2016:Survey}), and as such, the particular challenges faced by buses have been overlooked.

A particular behavior in bus drivers is that of using the height of the chassis when maneuvering.
Fig.~\ref{fig:bus} shows that the chassis ahead and behind the wheels of the bus, hereby referred to as overhangs, is quite elevated.
Bus drivers often allow the overhangs to sweep outside of the road limits and over curbs, in order to more easily maneuver the vehicle.
This work takes advantage of this behavior, planning collision free paths that can sweep outside of the road and over curbs.

The road-aligned vehicle model used in this work, is particularly suited for road driving, but it comes with the drawback of introducing distortions in the vehicle body, making the obstacle avoidance task challenging.
These distortions can often be ignored when considering short vehicles driving on low curvature roads, such as highways, but become critical when considering long vehicles driving on high curvature roads, as is the case in urban bus driving.
To deal with this, we propose novel vehicle body approximations, that capture the distortions of the road-aligned frame, and ensure safe and not excessively conservative obstacle avoidance.

\def\BusFontSize{\large}
\begin{figure}
  \centering
    \begin{tikzpicture}
	
\node[anchor=south west,inner sep=0] (image) at (0,0) {\includegraphics[width=0.99\linewidth]{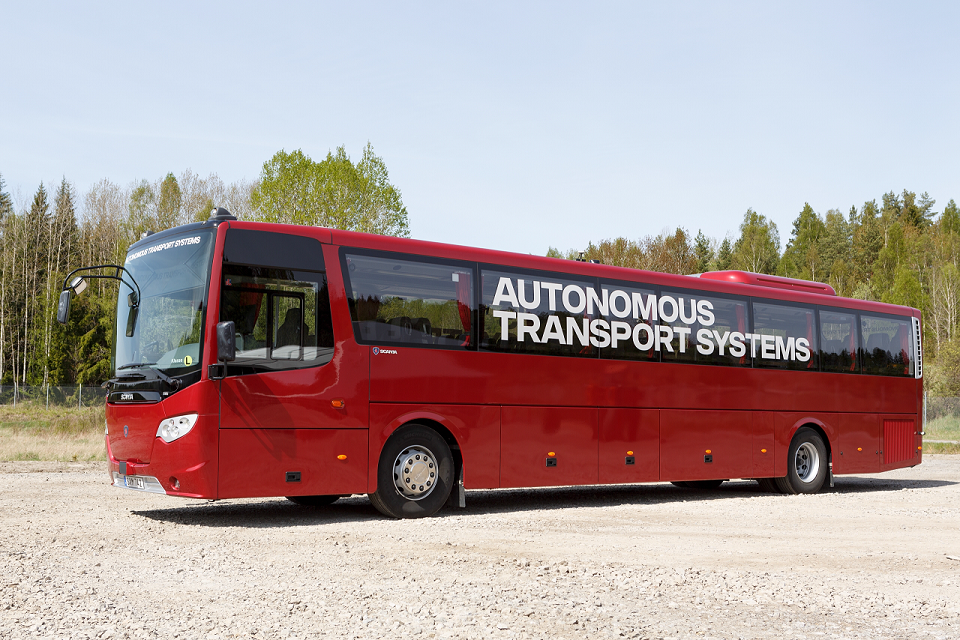}};
    \begin{scope}[x={(image.south east)},y={(image.north west)}]
	
	\def\NodeHeight{0.18}       
	\def\MarkerHeight{0.03} 
	\def\MarkerWidthSep{0.009}
	\def\TextVerticalOffset{0.013}
        
    \coordinate (A) at (0.12, \NodeHeight); 
    \coordinate (B) at (0.38, \NodeHeight); 

	\def\LineWidth{0.75mm}
    
	\def\FirstColor{green!70!black} 
	    
    \draw [\FirstColor,line width=\LineWidth] (A) -- (B);
    
    \node[black, below] at ($(A)!0.5!(B)+(0, \TextVerticalOffset)$) {\BusFontSize
\begin{tabular}{c}
Front \\
overhang \\
\end{tabular}    
    };
    \draw [\FirstColor,line width=\LineWidth] ($(A)+(0, \MarkerHeight)$) -- ($(A)-(0, \MarkerHeight)$);
    \draw [\FirstColor,line width=\LineWidth] ($(B)+(0, \MarkerHeight)$) -- ($(B)-(0, \MarkerHeight)$);
	
    \coordinate (C) at ($(B)+(\MarkerWidthSep, 0)$); 
    \coordinate (D) at (0.865, \NodeHeight); 	
	
	\def\SecondColor{red}    
	
	\draw [\SecondColor,line width=\LineWidth] (C) -- (D);
	\node[black, below] at ($(C)!0.5!(D)+(0, \TextVerticalOffset)$) {\BusFontSize Wheelbase};
	
    \draw [\SecondColor,line width=\LineWidth] ($(C)+(0, \MarkerHeight)$) -- ($(C)-(0, \MarkerHeight)$);
    \draw [\SecondColor,line width=\LineWidth] ($(D)+(0, \MarkerHeight)$) -- ($(D)-(0, \MarkerHeight)$);
    
    \coordinate (E) at ($(D)+(\MarkerWidthSep, 0)$); 
    \coordinate (F) at (0.96, \NodeHeight); 	
	
	\def\ThirdColor{\FirstColor}    
	
	\draw [\ThirdColor,line width=\LineWidth] (E) -- (F);
	\node[black, below] at ($(E)!0.1!(F)+(0, \TextVerticalOffset)$) {\BusFontSize
	\begin{tabular}{r}
Rear \\
overhang \\
\end{tabular}
};
    \draw [\ThirdColor,line width=\LineWidth] ($(E)+(0, \MarkerHeight)$) -- ($(E)-(0, \MarkerHeight)$);
    \draw [\ThirdColor,line width=\LineWidth] ($(F)+(0, \MarkerHeight)$) -- ($(F)-(0, \MarkerHeight)$);
    
    \end{scope}

	\end{tikzpicture}   
	\vspace{-6.0mm} 
  \caption{A prototype autonomous bus, the dimensions of which are used for the experimental results in this work.
  The distinct vehicle body, with its large and elevated overhangs, allows it to sweep over curbs and low height obstacles (courtesy of Scania CV AB). \vspace{-6.0mm} \label{fig:bus}}
\end{figure}
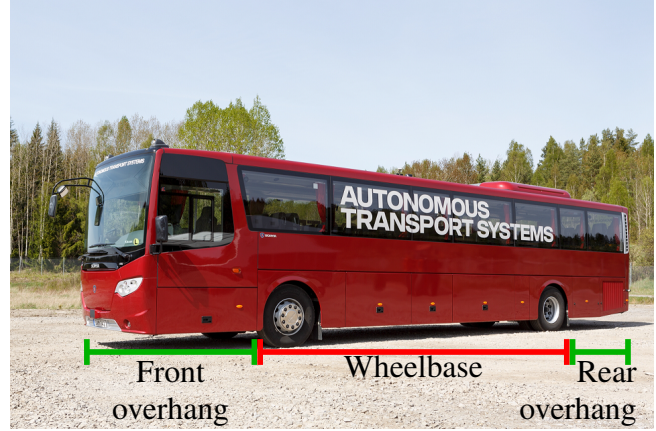

This work proposes a novel path planner that:
\begin{itemize}
\item tackles the challenging task of bus driving in urban environments, taking full advantage of the overhangs of buses to sweep over curbs and low height obstacles;
\item uses a new approximation technique for the distortions affecting the vehicle body and obstacles;
\item explicitly distinguishes between obstacle, sweepable, and drivable regions.
\end{itemize}

Section~\ref{sec:related_work} gives an overview of related work.
Section~\ref{sec:road-aligned_frame} introduces the road-aligned frame and the novel approximations for the vehicle body distortions.
Section~\ref{sec:problem_formulation} presents the challenges addressed in this work and a problem formulation.
Simulation results are presented in Section~\ref{sec:results}, and the final section concludes and lists future work directions.

\section{Related Work}
\label{sec:related_work}

In~\cite{Kuwata:2008:UrbanDrivingRRT}, the authors make use of rapidly-exploring random trees (RRT), proposing several algorithmic changes that make it suitable for autonomous road driving.
\cite{Evestedt:2015:SamplingRecovery} further expands on the RRT, targeting it to heavy-duty vehicles with limited retardation capabilities.
Both works disregard narrow roads, a common scenario for buses.

In~\cite{Ziegler:2009:SpatiotemporalLatticesOnRoad,McNaughton:2011:Conformal}, the solution space is discretized according to a state lattice.
Although efficient, this approach requires discretization, which invalidates infinitely many possible good solution paths, and can result in oscillatory behavior~\cite{Oliveira:2018:CombiningLattice}.
Furthermore, the discretization can make these planners fail in highly constrained environments.
To deal with this,~\cite{Fassbender:2016:HighlyConstrained} extends upon the A* search algorithm used in lattice search.

Optimization approaches can also be used, the main advantage being the smoothness of the solutions and a straightforward encoding of the vehicle model\cite{Schwarting:2018:Survey}.
A combined trajectory planning and control approach, formulated as a nonlinear model predictive control method is presented in~\cite{Gotte:2016:ARealTimeCapable}.
Simulation results show its effectiveness even in challenging emergency maneuvers.

Trajectory planning can also be formulated as sequential linear programming, as in~\cite{Plessen:2017:DimensionConstraintsSLP}, which targets the case of passenger vehicles.
However, the proposed solution cannot be applied to buses.
Long vehicles, together with sharp turns, result in unmodeled distortion effects that invalidate the collision avoidance constraints.
Thus,~\cite{Lima:2017:Overhang} introduces approximations that seek to ensure collision free paths.
Although safe, the approximations are too conservative, and the planner might fail in highly constrained environments.
\cite{Lima:2017:Overhang} is the first work that, to the best of our knowledge, addresses minimization of the bus overhang exiting the road.

In this work, we build upon~\cite{Lima:2017:Overhang}, and introduce a novel approximation method for vehicle body distortions, developing a path planner that guarantees safe collision free solutions, without conservative approximations.

\section{Modeling}
\label{sec:road-aligned_frame}

We introduce the road-aligned vehicle model used and propose a new approximation for the vehicle body, which deals with distortions introduced by the road-aligned frame.

\subsection{Vehicle model}
\label{subsec:vehicle_model}

We describe the vehicle state evolution using the space-based road-aligned vehicle model used in~\cite{Lima:2017:Overhang}.
The model describes the vehicle state using a frame that moves along a reference path.
This model is chosen since it allows for the convex formulation of common on-road optimization objectives and is independent of time.

\begin{figure}
  \centering
      \begin{tikzpicture}[scale=0.8]   
      \input{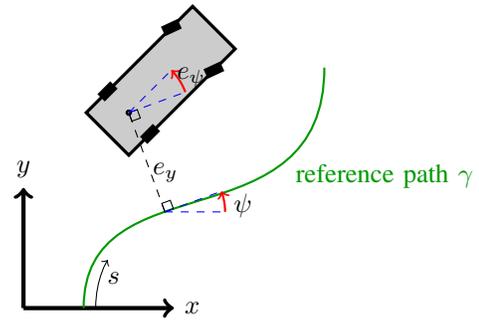}

\def\ReferenceOriginX{-1.0}
\def\ReferenceOriginY{-0.0}

\draw[->,ultra thick] (\ReferenceOriginX,\ReferenceOriginY)--(1.5,\ReferenceOriginY) node[right]{$x$};
\draw[->,ultra thick] (\ReferenceOriginX,\ReferenceOriginY)--(\ReferenceOriginX,2) node[above]{$y$};

\def\ReferencePathColor{green!60!black}

\draw[thick, color=\ReferencePathColor] (0,0) .. controls (0,2.5) and (4,1) .. (4,4);
\node[color=\ReferencePathColor] at (5.0,2.2) {reference path $\gamma$};

\draw [->] (0.2,0) .. controls (0.2,0.5) and (0.4,0.8) .. (0.4, 0.8);

\node[] at (0.5,0.5) {$s$};

\def\PathLocationX{1.35}
\def\PathLocationY{1.6}
\def\PathLocationO{20}

\def\VehicleX{1.0}
\def\VehicleX{0.75}
\def\VehicleY{3.25}
\def\VehicleO{45}
\def\VehicleHalfWidth{0.5 }
\def\VehicleFrontLength{2.0}
\def\VehicleRearLength{0.5}

\DrawVehicle{\VehicleX}{\VehicleY}{\VehicleO}{\VehicleHalfWidth}{\VehicleFrontLength}{\VehicleRearLength}

\draw[black,thick] ($(\VehicleX, \VehicleY)$) circle (1pt);
    
\draw[dashed] (\PathLocationX, \PathLocationY) -- ($(\VehicleX, \VehicleY)$);  

\node[] at ($(\PathLocationX, \PathLocationY)!0.4!(\VehicleX, \VehicleY) + (0.25, 0)$) {$e_y$};  

\draw[black, rotate around={\PathLocationO:(\PathLocationX, \PathLocationY)}] (\PathLocationX, \PathLocationY) rectangle ($(\PathLocationX, \PathLocationY)+(0.15, 0.15)$);

\draw[black, rotate around={\PathLocationO:(\VehicleX, \VehicleY)}] (\VehicleX, \VehicleY) rectangle ($(\VehicleX, \VehicleY)+(0.15, -0.15)$);

\draw[dashed, blue] (\PathLocationX, \PathLocationY) -- ($(\PathLocationX, \PathLocationY) + ({cos(\PathLocationO)}, {sin(\PathLocationO)})$);

\draw[dashed, blue] (\PathLocationX, \PathLocationY) -- ($(\PathLocationX, \PathLocationY) + (1, 0)$);

\draw [->, red,thick,domain=0:\PathLocationO] plot ({\PathLocationX+cos(\x)}, {\PathLocationY+sin(\x)});
\node[] at ($({\PathLocationX+1.3*cos(5)}, {\PathLocationY+1.3*sin(5)})$) {$\psi$};

\draw[dashed, blue] (\VehicleX, \VehicleY) -- ($(\VehicleX, \VehicleY) + ({cos(\PathLocationO)}, {sin(\PathLocationO)})$);

\draw[dashed, blue] (\VehicleX, \VehicleY) -- ($(\VehicleX, \VehicleY) + ({cos(\VehicleO)}, {sin(\VehicleO)})$);

\draw [->, red, thick, domain=\PathLocationO:\VehicleO] plot ({\VehicleX+cos(\x)}, {\VehicleY+sin(\x)});

\node[] at ($({\VehicleX+1.2*cos(30)}, {\VehicleY+1.2*sin(30)})$) {$e_\psi$};
    \end{tikzpicture}
  \caption{
  Global and road-aligned frames.
  Vehicle states $(s, e_y, e_\psi)$ on the road-aligned frame, are defined w.r.t. reference path $\gamma$. \label{fig:road_aligned} \vspace{-3.0mm} }
\end{figure}

As shown in Fig.~\ref{fig:road_aligned}, the road-aligned vehicle states are given by $(s, e_y, e_\psi)$ corresponding to the distance along the reference path $s$, the lateral displacement $e_y$, and the orientation difference $e_\psi$ between the vehicle heading and the path heading $\psi$.
These states are defined w.r.t. a reference path $\gamma$.
The vehicle is controlled by input $u$ corresponding to the vehicle curvature, which is related to the steering wheel angle $\phi$ as $u = \tan \left( \phi \right)/L$, where $L$ is the wheelbase length.

The reference path is uniformly discretized across its length every $\DiscretizationSamplingDistance$, so that $\{s_i\}^{N}_{i=0}$, where $s_i = i\DiscretizationSamplingDistance$.
To obtain a linear system, the vehicle model is linearized around reference states given by $\linearizationReferenceVariablesS = \{\bar{s}_{i}\}_{i=0}^{N}$, $\linearizationReferenceVariablesEy = \{\bar{e}_{y,i}\}_{i=0}^{N}$, $\linearizationReferenceVariablesEpsi = \{\bar{e}_{\psi,i}\}_{i=0}^{N}$, and $\linearizationReferenceVariablesU = \{\bar{u}_{i}\}_{i=0}^{N}$.
This results in a linear system of the form $z_{i+1} = A_i z_{i} + B_i u_i + G_i$, where $z_i = [e_{y,i}, e_{\psi,i}]^T$.
The reader is referred to~\cite{Lima:2017:Overhang} for further details about the formulation of the space-based road-aligned vehicle model, which are skipped for the sake of brevity.

\subsection{Conversion to road-aligned frame}
\label{subsec:conversion_geometry}

\begin{figure}
  \centering
       \resizebox{0.85\columnwidth}{!}{\def\tikzLegendSize{ \large} 
%
%
\definecolor{mycolor1}{rgb}{0.49804,0.72157,0.87059}%
\begin{tikzpicture}

\begin{axis}[%
width=3.566in,
height=3.566in,
at={(1.236in,0.481in)},
scale only axis,
xmin=1.2167,
xmax=10.1117,
xtick={3, 5, 7, 9},
xlabel style={font=\color{white!15!black}},
xlabel={$x$ [m]},
ymin=1.4925,
ymax=10.3875,
ytick={3, 5, 7, 9},
ylabel style={font=\color{white!15!black}},
ylabel={$y$ [m]},
axis background/.style={fill=white},
legend style={at={(0.03,0.97)}, anchor=north west, legend cell align=left, align=left, draw=white!15!black},
xlabel style={at={(axis cs:6,2)},font=\tikzLegendSize},ylabel style={at={(axis cs:2,6)},font=\tikzLegendSize},legend style={font=\tikzLegendSize},
]
\addplot [color=black, dashed]
  table[row sep=crcr]{%
0	0\\
0.00622370841970377	0.249896470390208\\
0.0249292568182811	0.499169579487612\\
0.056069891066397	0.747196274338393\\
0.0995677757961597	0.993356617375231\\
0.155314188949006	1.23703533593836\\
0.223169793524223	1.47762336013681\\
0.30296498584889	1.71451934520588\\
0.394500319498728	1.94713117455596\\
0.497547003810292	2.17487743975557\\
0.61184747573848	2.39718889374968\\
0.737116043630004	2.61350987368085\\
0.873039601303748	2.82329968975706\\
1.01927841065315	3.02603397669454\\
1.17546695081453	3.22120600435791\\
1.3412148317789	3.40832794432166\\
1.5161077701637	3.58693208918708\\
1.69970862470542	3.75657202160727\\
1.89155848888521	3.91682373009793\\
2.09117783795607	4.06728666884532\\
2.29806772750509	4.20758475886215\\
2.51171104055457	4.33736732798922\\
2.73157378008517	4.45630998739315\\
2.95710640375035	4.56411544236958\\
3.18774519744609	4.66051423542507\\
3.42291368430249	4.74526541978051\\
3.6620240655758	4.81815716161254\\
3.90447868983912	4.8790072695279\\
4.14967154679972	4.9276636499469\\
4.39698978200915	4.96400468725827\\
4.64581522868026	4.98793954779476\\
4.89552595278214	4.99940840687007\\
5.14549780755139	4.99838259830938\\
5.39510599353388	4.98486468609985\\
5.64372662025811	4.95888845798202\\
5.89073826563637	4.92051884099803\\
6.13552352919641	4.86985173920782\\
6.37747057526103	4.80701379397896\\
6.61597466221864	4.73216206744915\\
6.85043965406239	4.64548364995268\\
7.08027951041966	4.54719519239206\\
7.30491975134785	4.43754236472354\\
7.52379889323501	4.3167992419102\\
7.73636985221648	4.18526761887729\\
7.94210131159963	4.04327625618203\\
8.14047904987886	3.89118005828349\\
8.33100722602162	3.72935918646624\\
8.5132096188127	3.55821810863515\\
8.68663081715929	3.37818458835626\\
8.85083735838153	3.1897086156706\\
9.0054188116435	2.9932612823535\\
9.14998880381649	2.78933360443047\\
9.28418598521067	2.57843529489283\\
9.4076749327611	2.36109348968071\\
9.5201469884107	2.1378514301177\\
9.62132103059465	1.90926710509051\\
9.71094417689791	1.67591185636728\\
9.78879241612957	1.43836895054076\\
9.85467116823414	1.19723212116556\\
9.90841577064032	0.95310408473349\\
};
\addlegendentry{Reference path}

\addplot[area legend, draw=black, fill=mycolor1]
table[row sep=crcr] {%
x	y\\
9.4995290016229	7.90048345569412\\
9.49952238985379	6.10048345570627\\
1.99952238990439	6.10051100474422\\
1.99952900167349	7.90051100473207\\
}--cycle;
\addlegendentry{Vehicle body}

\addplot [color=white, draw=none, mark=*, mark options={solid, fill=red, red}]
  table[row sep=crcr]{%
1.99952238990439	6.10051100474422\\
3.87452238989174	6.10050411748473\\
5.74952238987909	6.10049723022524\\
7.62452238986644	6.10049034296575\\
9.49952238985379	6.10048345570627\\
};
\addlegendentry{Vehicle edge points}

\addplot [color=black, forget plot]
  table[row sep=crcr]{%
2.79327800234671	4.48718899991585\\
1.99952238990439	6.10051100474422\\
};
\addplot [color=black, forget plot]
  table[row sep=crcr]{%
3.15223077372285	4.66369086370598\\
2.97572890993271	5.02264363508212\\
2.61677613855658	4.84614177129199\\
2.79327800234671	4.48718899991585\\
3.15223077372285	4.66369086370598\\
};
\node[right, align=left]
at (axis cs:2,6.401) {\tikzLegendSize$\mathbf{\hat{p}}_{1}$};
\node[right, align=left]
at (axis cs:2.596,5.294) {\tikzLegendSize$\hat{e}_{y,1}$};
\addplot [color=black, draw=none, mark=*, mark options={solid, black}, forget plot]
  table[row sep=crcr]{%
2.79327800234671	4.48718899991585\\
};
\node[right, align=left]
at (axis cs:2.793,4.087) {\tikzLegendSize$\hat{s}_{1}$};
\addplot [color=black, forget plot]
  table[row sep=crcr]{%
4.09257493970575	4.91746856544083\\
3.87452238989174	6.10050411748473\\
};
\addplot [color=black, forget plot]
  table[row sep=crcr]{%
4.48593968171894	4.99002323269869\\
4.41338501446108	5.38338797471188\\
4.02002027244789	5.31083330745402\\
4.09257493970575	4.91746856544083\\
4.48593968171894	4.99002323269869\\
};
\node[right, align=left]
at (axis cs:3.875,6.401) {\tikzLegendSize$\mathbf{\hat{p}}_{2}$};
\node[right, align=left]
at (axis cs:4.184,5.509) {\tikzLegendSize$\hat{e}_{y,2}$};
\addplot [color=black, draw=none, mark=*, mark options={solid, black}, forget plot]
  table[row sep=crcr]{%
4.09257493970575	4.91746856544083\\
};
\node[right, align=left]
at (axis cs:4.093,4.517) {\tikzLegendSize$\hat{s}_{2}$};
\addplot [color=black, forget plot]
  table[row sep=crcr]{%
5.60999495568674	4.96315112787907\\
5.74952238987909	6.10049723022524\\
};
\addplot [color=black, forget plot]
  table[row sep=crcr]{%
6.00700216331053	4.91431182920723\\
6.05584146198236	5.31131903683103\\
5.65883425435857	5.36015833550286\\
5.60999495568674	4.96315112787907\\
6.00700216331053	4.91431182920723\\
};
\node[right, align=left]
at (axis cs:5.75,6.4) {\tikzLegendSize$\mathbf{\hat{p}}_{3}$};
\node[right, align=left]
at (axis cs:5.88,5.532) {\tikzLegendSize$\hat{e}_{y,3}$};
\addplot [color=black, draw=none, mark=*, mark options={solid, black}, forget plot]
  table[row sep=crcr]{%
5.60999495568674	4.96315112787907\\
};
\node[right, align=left]
at (axis cs:5.61,4.563) {\tikzLegendSize$\hat{s}_{3}$};
\addplot [color=black, forget plot]
  table[row sep=crcr]{%
6.9760831584733	4.59343973052818\\
7.62452238986644	6.10049034296575\\
};
\addplot [color=black, forget plot]
  table[row sep=crcr]{%
7.34350252580062	4.43531633405318\\
7.50162592227563	4.80273570138049\\
7.13420655494831	4.9608590978555\\
6.9760831584733	4.59343973052818\\
7.34350252580062	4.43531633405318\\
};
\node[right, align=left]
at (axis cs:7.625,6.4) {\tikzLegendSize$\mathbf{\hat{p}}_{4}$};
\node[right, align=left]
at (axis cs:7.5,5.347) {\tikzLegendSize$\hat{e}_{y,4}$};
\addplot [color=black, draw=none, mark=*, mark options={solid, black}, forget plot]
  table[row sep=crcr]{%
6.9760831584733	4.59343973052818\\
};
\node[right, align=left]
at (axis cs:6.976,4.193) {\tikzLegendSize$\hat{s}_{4}$};
\addplot [color=black, forget plot]
  table[row sep=crcr]{%
7.96791464897843	4.02436414044811\\
9.49952238985379	6.10048345570627\\
};
\addplot [color=black, forget plot]
  table[row sep=crcr]{%
8.28980003475096	3.78689877786638\\
8.52726539733268	4.10878416363891\\
8.20538001156016	4.34624952622064\\
7.96791464897843	4.02436414044811\\
8.28980003475096	3.78689877786638\\
};
\node[right, align=left]
at (axis cs:9.5,6.4) {\tikzLegendSize$\mathbf{\hat{p}}_{5}$};
\node[right, align=left]
at (axis cs:8.934,5.062) {\tikzLegendSize$\hat{e}_{y,5}$};
\addplot [color=black, draw=none, mark=*, mark options={solid, black}, forget plot]
  table[row sep=crcr]{%
7.96791464897843	4.02436414044811\\
};
\node[right, align=left]
at (axis cs:7.968,3.624) {\tikzLegendSize$\hat{s}_{5}$};
\end{axis}
\end{tikzpicture}%
 }
       \resizebox{0.85\columnwidth}{!}{\def\tikzLegendSize{ \Large} 
%
%
\definecolor{mycolor1}{rgb}{0.49804,0.72157,0.87059}%
\begin{tikzpicture}

\begin{axis}[%
width=4.523in,
height=3.333in,
at={(0.759in,0.597in)},
scale only axis,
xmin=4.569,
xmax=12.032,
xtick={ 5,  7,  9, 11},
xlabel style={font=\color{white!15!black}},
xlabel={$s$ [m]},
ymin=-1,
ymax=4.5,
ytick={0, 1, 3, 4},
ylabel style={font=\color{white!15!black}},
ylabel={$e_y$ [m]},
axis background/.style={fill=white},
legend style={at={(0.03,0.97)}, anchor=north west, legend cell align=left, align=left, draw=white!15!black},
xlabel style={at={(axis cs:8,-0.7)},font=\tikzLegendSize},ylabel style={at={(axis cs:5,2)},font=\tikzLegendSize},legend style={font=\tikzLegendSize},
]
\addplot [color=black, dashed]
  table[row sep=crcr]{%
0	0\\
20	0\\
};
\addlegendentry{Reference path}

\addplot[area legend, draw=black, fill=mycolor1]
table[row sep=crcr] {%
x	y\\
10.443	4.09150702317297\\
10.469	4.00930812839043\\
10.495	3.92735783141649\\
10.522	3.84566283805876\\
10.55	3.76423029496055\\
10.578	3.68306778715557\\
10.607	3.60218286579484\\
10.636	3.52158340355127\\
10.665	3.44127763988632\\
10.695	3.36127401346149\\
10.726	3.28158115436786\\
10.758	3.20220825864124\\
10.79	3.12316470550716\\
10.822	3.04446019599385\\
10.855	2.96610480414164\\
10.889	2.88810878790358\\
10.924	2.810482959008\\
10.959	2.73323864780239\\
10.995	2.6563872227195\\
11.032	2.5799406355762\\
11.032	2.5799406355762\\
10.816	2.3524880332129\\
10.586	2.13962862402556\\
10.344	1.9427044124123\\
10.087	1.76310782510729\\
9.817	1.60225308917417\\
9.535	1.4615398450754\\
9.241	1.34230901319583\\
8.937	1.24579093431072\\
8.624	1.17305118864454\\
8.306	1.12493666516357\\
7.983	1.10203034879434\\
7.66	1.10461554813585\\
7.338	1.13266042501241\\
7.021	1.1858185460543\\
6.71	1.2634504905417\\
6.407	1.36466093573868\\
6.116	1.48834626285736\\
5.836	1.63324968149132\\
5.569	1.79801442250809\\
5.569	1.79801442250809\\
5.599	1.88315078110101\\
5.629	1.9685349454826\\
5.658	2.05415794237861\\
5.686	2.14001112909445\\
5.713	2.22608644033759\\
5.74	2.31237581239719\\
5.767	2.39887187854654\\
5.792	2.48556749011353\\
5.817	2.57245578000283\\
5.842	2.65953011461116\\
5.866	2.74678433555349\\
5.889	2.83421239869486\\
5.912	2.92180849923219\\
5.934	3.00956725324133\\
5.956	3.09748316760722\\
5.978	3.18555124452474\\
5.999	3.27376673211083\\
6.019	3.36212495594404\\
6.039	3.45062135639094\\
6.039	3.45062135639094\\
6.261	3.31865386628809\\
6.49	3.20357464599044\\
6.724	3.10610316382496\\
6.964	3.02688058309183\\
7.208	2.96645156359499\\
7.455	2.92524597648935\\
7.704	2.90356447083262\\
7.954	2.90156774029547\\
8.203	2.91927076917969\\
8.451	2.95654192642701\\
8.695	3.01310837210136\\
8.936	3.08856500094285\\
9.172	3.18238944950294\\
9.402	3.29395835775175\\
9.625	3.42256654920534\\
9.841	3.56744676223655\\
10.05	3.72778867097807\\
10.25	3.90275693932041\\
10.443	4.09150702317297\\
}--cycle;
\addlegendentry{Vehicle body}

\addplot [color=black, forget plot]
  table[row sep=crcr]{%
5.569	0\\
5.569	1.79801442250809\\
};
\addplot [color=black, forget plot]
  table[row sep=crcr]{%
5.569	0.25\\
5.819	0.25\\
5.819	0\\
};
\addplot [color=red, draw=none, mark=*, mark options={solid, fill=red, red}, forget plot]
  table[row sep=crcr]{%
5.569	1.79801442250809\\
};
\node[right, align=left]
at (axis cs:5.669,1.798) {\tikzLegendSize$\mathbf{\hat{p}}_{1}$};
\node[right, align=left]
at (axis cs:5.569,0.5) {\tikzLegendSize$\hat{e}_{y,1}$};
\addplot [color=black, draw=none, mark=*, mark options={solid, black}, forget plot]
  table[row sep=crcr]{%
5.569	0\\
};
\node[right, align=left]
at (axis cs:5.569,-0.3) {\tikzLegendSize$\hat{s}_{1}$};
\addplot [color=black, forget plot]
  table[row sep=crcr]{%
6.942	0\\
6.942	1.20296302182578\\
};
\addplot [color=black, forget plot]
  table[row sep=crcr]{%
6.942	0.25\\
7.192	0.25\\
7.192	0\\
};
\addplot [color=red, draw=none, mark=*, mark options={solid, fill=red, red}, forget plot]
  table[row sep=crcr]{%
6.942	1.20296302182578\\
};
\node[right, align=left]
at (axis cs:7.042,1.203) {\tikzLegendSize$\mathbf{\hat{p}}_{2}$};
\node[right, align=left]
at (axis cs:6.942,0.5) {\tikzLegendSize$\hat{e}_{y,2}$};
\addplot [color=black, draw=none, mark=*, mark options={solid, black}, forget plot]
  table[row sep=crcr]{%
6.942	0\\
};
\node[right, align=left]
at (axis cs:6.942,-0.3) {\tikzLegendSize$\hat{s}_{2}$};
\addplot [color=black, forget plot]
  table[row sep=crcr]{%
8.466	0\\
8.466	1.14587262006488\\
};
\addplot [color=black, forget plot]
  table[row sep=crcr]{%
8.466	0.25\\
8.716	0.25\\
8.716	0\\
};
\addplot [color=red, draw=none, mark=*, mark options={solid, fill=red, red}, forget plot]
  table[row sep=crcr]{%
8.466	1.14587262006488\\
};
\node[right, align=left]
at (axis cs:8.566,1.146) {\tikzLegendSize$\mathbf{\hat{p}}_{3}$};
\node[right, align=left]
at (axis cs:8.466,0.5) {\tikzLegendSize$\hat{e}_{y,3}$};
\addplot [color=black, draw=none, mark=*, mark options={solid, black}, forget plot]
  table[row sep=crcr]{%
8.466	0\\
};
\node[right, align=left]
at (axis cs:8.466,-0.3) {\tikzLegendSize$\hat{s}_{3}$};
\addplot [color=black, forget plot]
  table[row sep=crcr]{%
9.886	0\\
9.886	1.64063249549013\\
};
\addplot [color=black, forget plot]
  table[row sep=crcr]{%
9.886	0.25\\
10.136	0.25\\
10.136	0\\
};
\addplot [color=red, draw=none, mark=*, mark options={solid, fill=red, red}, forget plot]
  table[row sep=crcr]{%
9.886	1.64063249549013\\
};
\node[right, align=left]
at (axis cs:9.986,1.641) {\tikzLegendSize$\mathbf{\hat{p}}_{4}$};
\node[right, align=left]
at (axis cs:9.886,0.5) {\tikzLegendSize$\hat{e}_{y,4}$};
\addplot [color=black, draw=none, mark=*, mark options={solid, black}, forget plot]
  table[row sep=crcr]{%
9.886	0\\
};
\node[right, align=left]
at (axis cs:9.886,-0.3) {\tikzLegendSize$\hat{s}_{4}$};
\addplot [color=black, forget plot]
  table[row sep=crcr]{%
11.032	0\\
11.032	2.5799406355762\\
};
\addplot [color=black, forget plot]
  table[row sep=crcr]{%
11.032	0.25\\
11.282	0.25\\
11.282	0\\
};
\addplot [color=white, draw=none, mark=*, mark options={solid, fill=red, red}]
  table[row sep=crcr]{%
11.032	2.5799406355762\\
};
\addlegendentry{Vehicle edge points}

\node[right, align=left]
at (axis cs:11.132,2.58) {\tikzLegendSize$\mathbf{\hat{p}}_{5}$};
\node[right, align=left]
at (axis cs:11.032,0.5) {\tikzLegendSize$\hat{e}_{y,5}$};
\addplot [color=black, draw=none, mark=*, mark options={solid, black}, forget plot]
  table[row sep=crcr]{%
11.032	0\\
};
\node[right, align=left]
at (axis cs:11.032,-0.3) {\tikzLegendSize$\hat{s}_{5}$};
\end{axis}
\end{tikzpicture}%
 }
  \caption{
  The vehicle body in the road aligned frame (bottom), can be converted to the cartesian frame (top) using geometric method $\geometricTransform$. $\geometricTransform$ finds the normal projection of the vehicle body in the reference path, shown in the top figure. The resulting vehicle body in the road aligned frame becomes distorted, as shown in the bottom figure. \label{fig:projection_and_distortion} \vspace{-5.0mm} }
\end{figure}
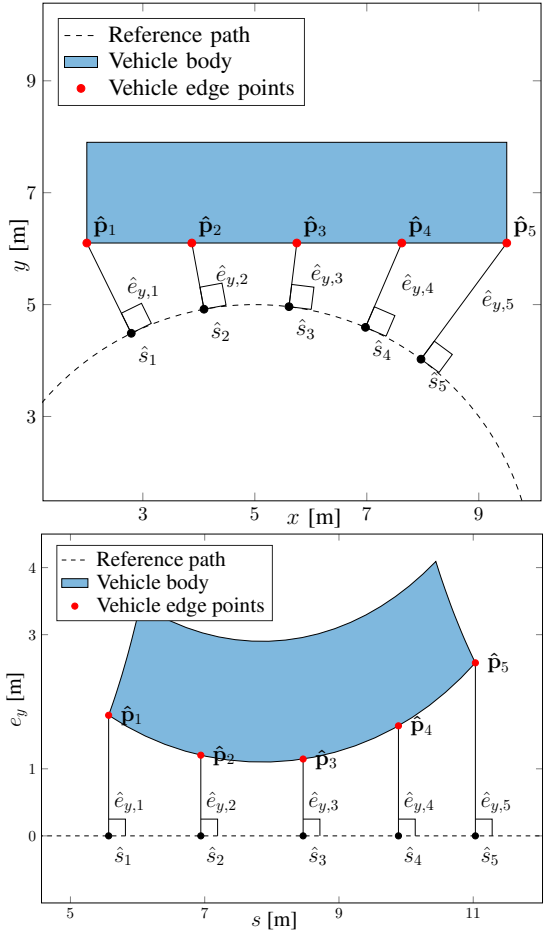

A cartesian position $\CartesianPosition \in \realNumbers^2$, can be converted to the road-aligned frame using a geometric algorithm.
We define the reference path $\gamma$ as the map $\gamma:\linearizationReferenceVariablesS \rightarrow (x, y) \in \realNumbers^2$, where the domain is the discretized path length.
One can convert $\CartesianPositionConverted$ to the road-aligned frame, by finding the location $\gamma(\SConverted)$, in which the normal to the path is pointing towards $\CartesianPositionConverted$.
$\EYConverted$ is then $|| \gamma(\SConverted) - \CartesianPositionConverted ||$.
We define this conversion as $\geometricTransform : \CartesianPosition \in \realNumbers^2 \rightarrow (s, e_y) \in \linearizationReferenceVariablesS\times\realNumbers$.
$\geometricTransform$ is useful when one needs to evaluate the vehicle body in the road-aligned frame.
Fig.~\ref{fig:projection_and_distortion} illustrates the results of $\geometricTransform$, when converting points along the vehicle edge.

\subsection{Distortions in the road-aligned frame}

When planning a path it is necessary to take into account the vehicle body and check it against obstacles.
When using the road-aligned model, it is necessary to account for the heavy distortion of objects due to transformation $\geometricTransform$, shown in Fig.~\ref{fig:projection_and_distortion}.
Since $\geometricTransform$ is obtained via a geometric algorithm, an analytical approximation of it is of interest, in order to be able to use optimization algorithms.
In the following, we derive such an approximation.

Given a road aligned vehicle state $(s, e_y, e_\psi)$ and the corresponding cartesian state $(x, y, \psi)$, corresponding to position and orientation, one can compute the first order Taylor expansion for a point along the vehicle edge.
Assuming a vehicle body edge point located at position $\CartesianPositionConverted$, one can get the corresponding road aligned coordinates as $(\SConverted, \EYConverted)=\geometricTransform(\CartesianPositionConverted)$ (see Fig.~\ref{fig:projection_and_distortion}). 
Assuming a fixed $\SConverted$, the first order Taylor expansion w.r.t. $e_y$ and $e_\psi$, around linearization point $(\bar{s}, \bar{e}_y, \bar{e}_\psi)$, is computed: 
\begin{equation}
\EYConverted = \geometricTransformEY(\CartesianPositionConverted) + \frac{\partial \geometricTransformEY(\CartesianPositionConverted)}{\partial e_y}( e_{y} - \bar{e}_y ) + \frac{\partial \geometricTransformEY(\CartesianPositionConverted)}{\partial e_\psi}( e_\psi - \bar{e}_\psi ),
\label{eq:taylor_approximation}
\end{equation}
where $\geometricTransformEY$ is $\geometricTransform$ with a co-domain corresponding to the lateral displacement $e_y$ only.
Note that $\CartesianPosition'$ depends on vehicle states $e_y$ and $e_\psi$, as they determine the vehicle position and orientation, and by consequence, the location of points on the vehicle body.

The partial derivatives can be approximated via a finite difference formula.
However this requires numerous calls to the geometric method $\geometricTransform_{e_y}$, which is computationally expensive.
Instead, we propose an alternative approximation to the partial derivatives which is faster to compute.

\subsection{Arc-circle approximation}
\label{subsec:arc-circle-approximation}

\def\tikzLegendSize{ \large }
\begin{figure}
  \centering
      \resizebox {0.85\columnwidth} {!} { \input{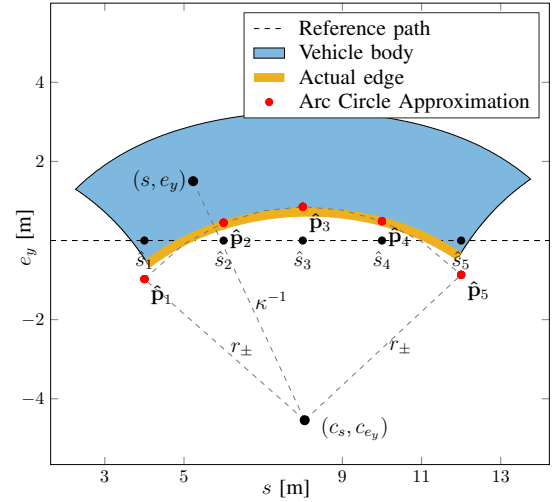} }
  \caption{Distorted vehicle edges can be approximated by arc-circles. \label{fig:approximation_distortion_arc_circles}  \vspace{-10.0mm} }
\end{figure}

In the road-aligned coordinate frame, the edges of the vehicle body are distorted to curves that resemble arc-circles, as seen in Fig.~\ref{fig:approximation_distortion_arc_circles}.
We exploit this insight to formulate an approximation to the partial derivatives in Eq.~\eqref{eq:taylor_approximation}.

During experiments, it was observed that the edges can be approximated by an arc-circle with a radius similar to the inverse of the reference path curvature, $\kappa^{-1}$, evaluated at length $s$.
Moreover, the center of the arc-circle is located perpendicularly to the vehicle rear axle $(s, e_y, e_\psi)$.
We thus have for the center of the arc-circle (see Fig.~\ref{fig:approximation_distortion_arc_circles}):
\begin{equation}
\begin{aligned}
c_s &= s + \kappa^{\text{-}1}\sin e_\psi, \\
c_{e_y} &= e_y - \kappa^{\text{-}1}\cos e_\psi.
\end{aligned}
\label{eq:arc-circle-1}
\end{equation}

Depending on the edge to be considered, left or right, the arc radius $\arcCircleRadius$ is equal to the inverse of the road curvature, plus or minus half the width $\omega$ of the vehicle, that is, $\arcCircleRadius = \kappa^{-1} \pm \omega/2$.
The expression of the circle to which the arc belongs to is then: \begin{equation}
\begin{aligned}
\left( \EYConverted - c_{e_y} \right)^2 + \left( \SConverted - c_{s} \right)^2  = \arcCircleRadius^2.
\end{aligned}
\label{eq:arc-circle-2}
\end{equation}
Assuming a constant $\SConverted$, and writing in order to $\EYConverted$, the previous expression becomes:
\begin{equation}
\EYConverted(e_y, e_\psi) = c_{e_y} + \sqrt{\arcCircleRadius^2 - (\SConverted - c_s)^2},
\label{eq:arc-circle-3}
\end{equation}
where $\EYConverted$ corresponds to the lateral offset of the edge, evaluated at length $\SConverted$.
We write $\EYConverted(e_y, e_\psi)$ to highlight the dependency on vehicle states $e_y$ and $e_\psi$.

In essence, this approximation assumes that the different points along the vehicle edge can be thought to belong to an arc-circle that is attached to the vehicle state $(e_y, e_\psi)$.
Relying on this dependency, we approximate the partial derivatives in equation~\eqref{eq:taylor_approximation} by the partial derivatives of $\EYConverted$:
\begin{equation}
\begin{aligned}
\frac{\partial \geometricTransformEY(\CartesianPositionConverted)}{\partial e_y} \approx \frac{\partial \EYConverted(e_y, e_\psi)}{\partial e_{y}}, \\ 
\frac{\partial \geometricTransformEY(\CartesianPositionConverted)}{\partial e_\psi} \approx \frac{\partial \EYConverted(e_y, e_\psi)}{\partial e_{\psi}}.
\end{aligned}
\end{equation}
Doing so, we skip the computationally expensive process of computing the partial derivatives of $\geometricTransformEY$ via finite differences.

The previous procedure gives us an expression for Eq.~\eqref{eq:taylor_approximation}.
The positional constraint $p_{e_y} \leq \EYConverted$, which forces a vehicle body edge point to be contained in a certain region, can then written, by reorganizing the terms in Eq.~\eqref{eq:taylor_approximation}, as:
\begin{equation}
p_{e_y} \leq P z + p.
\label{eq:vehicle_body_edge_position_constraint}
\end{equation}
Where $p_{e_y} \in \realNumbers$ is the position constraint (\textit{e.g.}, corresponding to the boundary of an obstacle), $P \in \realNumbers^2$ is a row vector for the terms associated with $z = [e_{y}, e_{\psi}]^T$, and $p \in \realNumbers$ is a scalar, encompassing all constant terms in Eq.~\eqref{eq:taylor_approximation}.

\section{Problem Formulation}
\label{sec:problem_formulation}

We introduce the objectives and constraints that path solutions must take into account.
Special attention is given to the challenges faced by buses, which must be dealt with by developing special constraints for the overhang parts.

\subsection{Driving objectives}

A goal in on-road driving is to drive as much as possible in the center of the lane.
Assuming that the reference path corresponds to the center of the lane, as is often the case in on-road driving, we define the optimization objective $\optObjCenterDriving$ to be the squared Euclidean norm of the lateral displacement, $\|(e_{y,0}, e_{y,1}, \ldots, e_{y,N})\|_2^2$.

Passenger comfort is also of importance, especially in buses.
Thus, we introduce the minimization objective $\optObjSmooth$ given by $\sum_{i=1}^{N\text{-}1} \left( u_i - u_{i-1} \right)^2$.
Minimizing $\optObjSmooth$ results in a smooth control input profile, \textit{i.e.} steering profile, which in turn results in more comfortable driving.

\subsection{Overhangs and environment classification}

Buses have relatively big overhangs (see Fig.~\ref{fig:bus}), when compared to other vehicles.
The overhangs allow the bus to have a large passenger capacity, while keeping the wheelbase small, which increases the turning radius.
Furthermore, the smaller wheelbase allows for a better load balance on the vehicle chassis.
Experienced drivers take advantage of the height of the overhangs, and use it to better maneuver the vehicle.
Often a driver maneuvers the bus in a way that allows the overhangs to sweep over curbs.

Planning approaches typically take into account the dimensions of the vehicle body and use it to compute collision free paths.
It is common to split the planning space using a binary classification into obstacle or obstacle free regions~\cite{LaValle.Planning2006}.
Buses suffer from such a classification scheme, as they do not allow sweeping over low height obstacles.

\begin{figure}
  \centering
    \resizebox{0.9\columnwidth}{!}{
    \begin{tikzpicture}
	\input{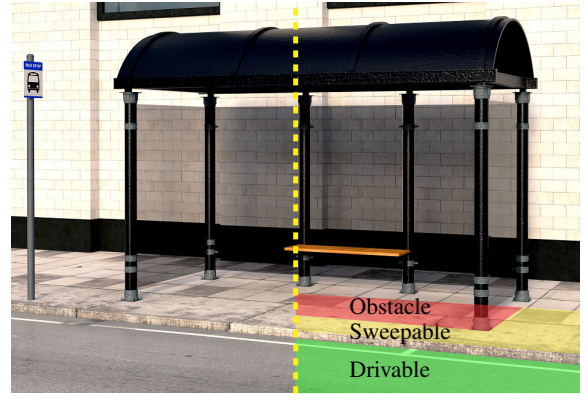}
	\end{tikzpicture}   
	}
  \caption{Example of a bus stop. On the right half of the image are overlayed the different region types. In red, yellow, and green are the obstacle, sweepable, and drivable regions.
  The vehicle body cannot enter the obstacle region in order to avoid collisions, however the overhangs are allowed to enter the sweepable region.\label{fig:bus_stop_regions} \vspace{-4.0mm} }
\end{figure}

To address this issue, we introduce a three-label approach, classifying the space into three different regions, as shown in Fig.~\ref{fig:bus_stop_regions}.
The obstacle region corresponds to obstacles that the vehicle body cannot collide with.
The sweepable region corresponds to obstacles of height lower than the overhangs, such as curbs, that can be swept over by the overhangs of the bus.
The drivable region corresponds to the road lane, where the wheels are allowed to be.

To formulate the obstacle constraints we make use of the arc-circle approximation introduced in Sec.~\ref{subsec:arc-circle-approximation}, and repeat it for $K$ equispaced points along the vehicle edge, for both edges.
Each point is then constrained, using Eq.~\eqref{eq:vehicle_body_edge_position_constraint}, to be inside the left or right obstacle region boundaries, depending on which vehicle edge is considered.
The obstacle constraints for all vehicle edge points can then be packed together as:
\begin{equation}
\optConstraintObsPosition \leq \optConstraintObsMatrix z_{i} + \optConstraintObsOffset, \; i \in [1, ..., N],
\label{eq:lp_cnst_obs}
\end{equation}
where $\optConstraintObsPosition \in \realNumbers^{2K}$, $\optConstraintObsMatrix \in \realNumbers^{2K \times 2}$, and $\optConstraintObsOffset \in \realNumbers^{2K}$.
The $2K$ rows of Eq.~\eqref{eq:lp_cnst_obs} correspond each to a positional constraint in the form of Eq.~\eqref{eq:vehicle_body_edge_position_constraint}.

Analogously, we limit the wheelbase to be inside the drivable region, by formulating the arc-circle approximation for $M$ equispaced points along the wheelbase edges.
The resulting drivable region constraints are:
\begin{equation}
\optConstraintRoadPosition \leq \optConstraintAxleMatrix z_{i} + \optConstraintAxleOffset, \; i \in [1, ..., N],
\label{eq:lp_cnst_wheel}
\end{equation}
where $\optConstraintRoadPosition \in \realNumbers^{2M}$, $\optConstraintAxleMatrix \in \realNumbers^{2M \times 2}$, and $\optConstraintAxleOffset \in \realNumbers^{2M}$.

To minimize overhangs entering the sweepable region, we first introduce optimization variable $\sigma$ corresponding to the amount of overhang exiting the drivable region.
Then, the arc-circle approximation is used for the four vehicle corner points.
Combining with the drivable region limits, together with the constraint that $\sigma$ must be non-negative, we get:
\begin{equation}
\begin{aligned}
\optConstraintOverhangRoadPosition &\leq \optConstraintOverhangMatrix z_{i} + \optConstraintOverhangOffset - \optConstraintOverhangSlack, \; i \in [1, ..., N], \\
\optConstraintOverhangSlack &\geq 0, \; i \in [1, ..., N],
\end{aligned}
\label{eq:lp_cnst_overhang}
\end{equation}
where $\optConstraintOverhangRoadPosition \in \realNumbers^{4}$, $\optConstraintOverhangMatrix \in \realNumbers^{4 \times 2}$, $\optConstraintOverhangOffset \in \realNumbers^{4}$, and $\optConstraintOverhangSlack \in \realNumbers^{4}$.

The optimization variable $\sigma$ is then penalized through objective $\optObjOverhang = \| (\mathbf{\sigma}_{1}^{r}, \mathbf{\sigma}_{2}^{r}, \ldots, \mathbf{\sigma}_{N}^{r}) \|_2^2$.
This minimization objective, together with the constraints defined previously, make $\sigma$ a measurement of the amount of overhang that exits the drivable region.
Thus, minimizing $\optObjOverhang$ results in reducing the amount of overhang exiting the road.

\subsection{System constraints}

We also define the state evolution constraints, corresponding to the discretized space-based road-aligned vehicle model introduced in Section~\ref{subsec:vehicle_model}:
\[
z_{i+1} = A_{i} z_{i} + B_{i}u_i + G_i, \; i \in [0, ..., N\text{-}1].
\]

Furthemore, it is necessary for the planned path to start from the current vehicle state, and with the current steering angle.
This originates constraints:
\[
z_0 = z_{\text{start}}, u_0 = u_{\text{start}}.
\]

Finally, we introduce constraints related to actuator limits, which are formulated as:
\begin{equation*}
\begin{aligned}
\optConstraintMaxU &\geq u_i \geq -\optConstraintMaxU, i \in [1, ..., N\text{-}1], \\
\optConstraintMaxUDot &\geq u_i - u_{i\text{-}1} \geq -\optConstraintMaxUDot, i \in [1, ..., N\text{-}1].
\end{aligned}
\end{equation*}
$\optConstraintMaxU$ and $\optConstraintMaxUDot$ are space-based limitations of the curvature that reflect magnitude and rate limits of the steering actuator.

\subsection{Sequential Quadratic Programming (SQP) formulation}

Combining all the optimization objectives and constraints mentioned before, one can formulate the following QP,~\cite{boyd2004convex}:

\begin{equation}
\begin{aligned} 
& \underset{\optimizationVariableU}{\text{min.}}
& & \optObjCenterDriving + \optObjSmooth + \optObjOverhang\label{eq:lp_problem}\\
& \text{~~s. t.}
& & z_{i+1} = A_{i} z_{i} + B_{i}u_i + G_i, \; i \in [0, ..., N\text{-}1],\\
&&& z_0 = z_{\text{start}}, u_0 = u_{\text{start}},\\
&&& \optConstraintObsPosition \leq \optConstraintObsMatrix z_{i} + \optConstraintObsOffset, \; i \in [1, ..., N],\\
&&& \optConstraintRoadPosition \leq \optConstraintAxleMatrix z_{i} + \optConstraintAxleOffset, \; i \in [1, ..., N],\\
&&& \optConstraintOverhangRoadPosition \leq \optConstraintOverhangMatrix z_{i} + \optConstraintOverhangOffset - \optConstraintOverhangSlack, \; i \in [1, ..., N],\\
&&& \optConstraintOverhangSlack \geq 0, \; i \in [1, ..., N],\\
&&& \optConstraintMaxU \geq u_i \geq -\optConstraintMaxU, i \in [1, ..., N\text{-}1]\\
&&& \optConstraintMaxUDot \geq u_i - u_{i\text{-}1} \geq -\optConstraintMaxUDot, i \in [1, ..., N\text{-}1]
\end{aligned} 
\end{equation}
With optimization variable $\optimizationVariableU$ corresponding to control inputs $(u_0, u_1, \ldots, u_{N\text{-}1})$.

The optimal inputs $\solutionVariableU$, and vehicle states  $\solutionVariablesEy$, $\solutionVariablesEpsi$, which are the solution to the optimization problem~\eqref{eq:lp_problem}, can be relatively far from the linearization references $\linearizationReferenceVariablesU$, $\linearizationReferenceVariablesEy$, $\linearizationReferenceVariablesEpsi$.
This means that the vehicle body approximations~\eqref{eq:lp_cnst_obs}, \eqref{eq:lp_cnst_wheel}, and \eqref{eq:lp_cnst_overhang} lose accuracy, possibly resulting in planned paths that violate the vehicle body constraints.

To overcome this problem, we use Sequential Quadratic Programming (SQP)~\cite{Plessen:2017:DimensionConstraintsSLP}.
In SQP, problem~\eqref{eq:lp_problem} is sequentially solved, and at each iteration, the previous solution becomes the linearization reference for the current QP.
Thus, we can guarantee that the final QP solution has an arbitrarily small distance to the linearization reference.
By setting the allowed distance to a small value, we can enforce the quality of our approximations.

As the succesive linearizations of the problem are solved, one gets that $\solutionVariablesEy \rightarrow \linearizationReferenceVariablesEy$ and $\solutionVariablesEpsi \rightarrow \linearizationReferenceVariablesEpsi$.
This in turn indicates that the first order Taylor expansion~\eqref{eq:taylor_approximation} converges to the constant term, \textit{i.e.}, $\EYConverted \rightarrow \geometricTransformEY(\CartesianPositionConverted)$, corresponding to the exact value of the edge location.
Thus, as SQP progresses along iterations, so does the approximation become more accurate.

\section{Results}
\label{sec:results}

\def\tikzLegendSize{ \LARGE }
\def\tikzTickSize{ \LARGE }
\def\tikzLabelSize{ \huge }
\begin{figure}
  \centering
      \resizebox {\columnwidth} {!} { \input{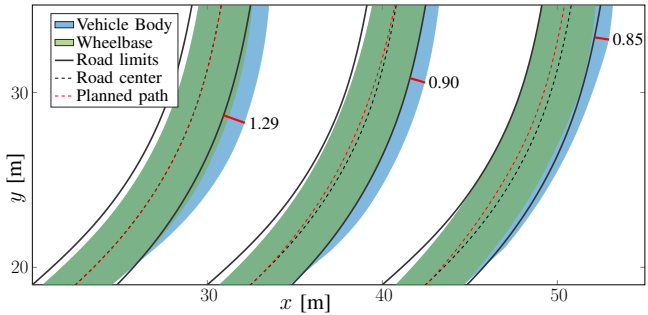} }
  \caption{The influence of wheelbase constraints and overhang minimization on the planned path. Disregarding wheelbase constraints and overhang minimization (left), considering wheelbase constraints only (center), and considering both wheelbase constraints and overhang minimization (right). The maximum amount (in meters) that the vehicle body exits the road is shown in red, and it decreases from $1.29$ (left) to $0.90$ (center) and finally to $0.85$ (right). \label{fig:wheelbase_overhang_influence} \vspace{-7.0mm} }
\end{figure}

\def\tikzLegendSize{ \large }
\begin{figure*}[t]
  \centering
      \resizebox {\textwidth} {!} { \input{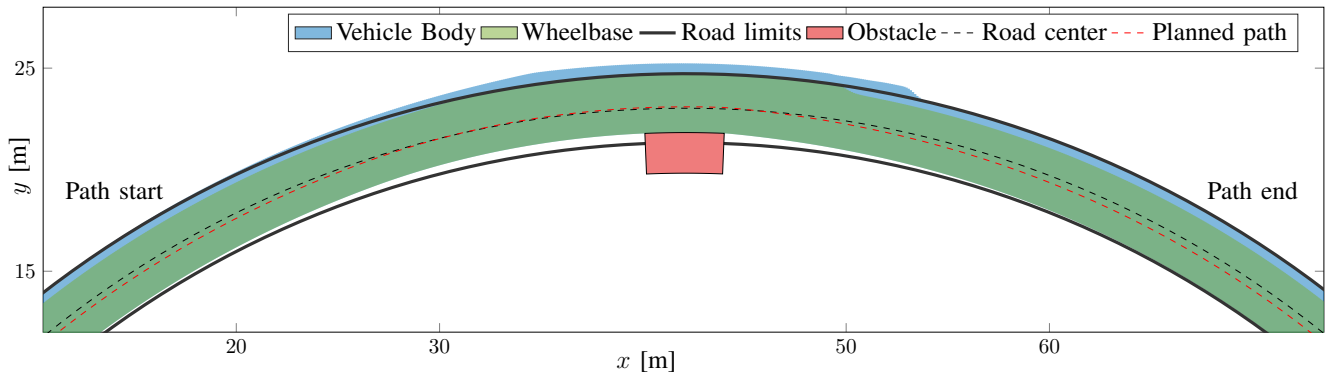} }
  \caption{Planned path on a road with an obstacle forcing the bus to drive through a passage with small obstacle clearance. \label{fig:thight_obstacles} \vspace{-2.0mm} }
\end{figure*}

\def\tikzLegendSize{ \Large }
\begin{figure}
  \centering
      \resizebox {0.9\columnwidth} {!} { \input{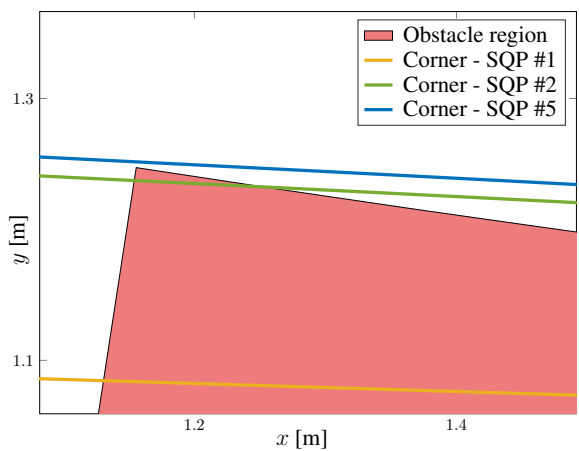} }
  \caption{Zoomed view of a selected problem instance. The curves represent the location of the front right corner of the bus, according to the planned path, at three different SQP iterations (SQP converged at iteration 5). At each iteration the distortion approximation becomes more accurate, and the planned paths converge to a solution avoiding the obstacle. \label{fig:successive_sqp} \vspace{-7.0mm} }
\end{figure}

\def\tikzLegendSize{ \large }

\subsection{Wheelbase constraints and overhang minimization}
\label{subsec:wheelbase_overhangs}

Fig.~\ref{fig:wheelbase_overhang_influence} shows the influence of wheelbase constraints \eqref{eq:lp_cnst_wheel} and optimization objective $\optObjOverhang$ on the planned paths.
If both $\optObjOverhang$ and \eqref{eq:lp_cnst_wheel} are disregarded, the vehicle follows the center of the road (Fig.~\ref{fig:wheelbase_overhang_influence} left).
Considering constraint \eqref{eq:lp_cnst_wheel} forces the vehicle to the inside of the turn in order to keep the wheelbase inside the road lane (Fig.~\ref{fig:wheelbase_overhang_influence} center).
By also minimizing $\optObjOverhang$ the planned path is further pushed to the inside of the turn (Fig.~\ref{fig:wheelbase_overhang_influence} right).

The maximum amount that the vehicle body exited the road is also measured, and it can be seen that it is greatly reduced from $1.29$ meters to $0.85$ meters once the constraints and optimization objective are added.
This results in less invasive maneuvers for vehicles on adjacent lanes.

\subsection{Highly constrained maneuver}
\label{subsec:obstacle_avoidance}

One of the biggest challenges that path planners face are highly constrained scenarios, where the solution must pass through small obstacle free regions~\cite{Fassbender:2016:HighlyConstrained}.
We set up a scenario where an obstacle on the road forms a passage with low clearance, as illustrated in Fig.~\ref{fig:thight_obstacles}.
One could imagine this to be a possible representation of a scenario in which there is a temporarily stopped vehicle on the side of the road.

Fig.~\ref{fig:thight_obstacles} shows that the path planner is able to find a collision free solution that makes the vehicle progress through the low clearance passage.
Furthermore, the planned path makes use of the sweepable region, allowing the overhangs to exit the road limits, while keeping the wheelbase contained in the drivable region corresponding to the road.
If the overhangs were not allowed to exit the road limits, as is the case with other planners, then it would not be possible to find a solution.
This illustrates the importance of allowing the overhangs to go over sweepable regions.

We note that the planned path takes the turn on the inside, except when avoiding the obstacle.
This is done to minimize the amount of overhang exiting the driving lane.

\begin{remark}
The proposed planner assumes that a reference path is already obtained.
The reference path usually corresponds to the road center, but can also be obtained by using a simplified path planner that determines if obstacles should be avoided by driving through the left or right of them.
\end{remark}

\subsection{Improvement of distortion approximations}
\label{subsec:improvement_distortion_approximation}

We present in Fig.~\ref{fig:successive_sqp} a zoomed-in version of a selected planning instance, in which an obstacle is present on the road.
The figure shows the position of the vehicle's front right corner, when following the planned path, for different iterations of the SQP.
In initial iterations the roughness of approximations makes the vehicle corner intersect the obstacle.
However, the SQP iterations improve the accuracy of the approximation, resulting in successful obstacle avoidance.

The results show that the proposed path planner is capable of reducing the amount of overhang exiting the road, resulting in safer driving.
Furthermore it is capable of dealing with highly constrained scenarios, where the bus can barely fit.
This is achieved making use of approximations which are precise, being both safe and not conservative.

\section{Conclusions}
\label{sec:conclusions}

We present a novel path planning framework targeted for buses driving in urban environments.
The problem is solved via SQP, benefiting from a successive accuracy improvement of the approximations.
The proposed approximations guarantee obstacle avoidance, without being conservative.

The path planner also exploits the special body characteristics of buses, namely the overhangs.
Using a new labeling approach, which takes into account low height structures that can be swept by the overhangs, the planner is able to plan paths otherwise impossible when considering a binary classification into obstacle or obstacle free regions.

The authors are confident that the proposed solution can be used in an online fashion, and plan to pursue this in future work.
The quality of the vehicle body distortion approximations should be further studied.
We believe that the approximation error can be bounded and taken into account, so that all solution paths are guaranteed collision free even during intermediate iterations of the SQP.
It is also of interest to tackle the problem of bus stop approach maneuvers, which require stopping at a desired position with high precision.
Moreover, the framework will be extended to other vehicle configurations, such as, \textit{e.g.}, articulated buses.

\bibliographystyle{IEEEtran}
{\tiny
\bibliography{references}
}



\end{document}